\documentclass[11pt,a4paper]{article}
\usepackage[hyperref]{emnlp2020}
\usepackage{times}
\usepackage{latexsym}
\usepackage{booktabs}

\usepackage{nopageno}
\usepackage{microtype}

\usepackage{multirow}
\usepackage{makecell}
\usepackage{graphics}
\usepackage{chngcntr}

\newcommand{\todo}[1]{\textcolor{red}{TODO: #1}}

\newcommand{\taen}{Ta $\rightarrow$ En }
\newcommand{\enta}{En $\rightarrow$ Ta }
\newcommand{\iuen}{Iu $\rightarrow$ En }
\newcommand{\eniu}{En $\rightarrow$ Iu }
\newcommand{\taennosp}{Ta $\rightarrow$ En}
\newcommand{\entanosp}{En $\rightarrow$ Ta}
\newcommand{\iuennosp}{Iu $\rightarrow$ En}
\newcommand{\eniunosp}{En $\rightarrow$ Iu}

\title{Facebook AI's WMT20 News Translation Task Submission}

\author{Peng-Jen Chen, Ann Lee, Changhan Wang, Naman Goyal \\
{\bf Angela Fan, Mary Williamson, Jiatao Gu} \\
Facebook AI \\
  \texttt{\{pipibjc,annl,changhan,namangoyal\}@fb.com} \\
  \texttt{\{angelafan,marywilliamson,jgu\}@fb.com} \\
  }

\date{}

\begin{document}
\maketitle
\begin{abstract}

This paper describes Facebook AI's submission to WMT20 shared news translation task. We focus on the low resource setting and participate in two language pairs, Tamil $\leftrightarrow$ English and Inuktitut $\leftrightarrow$ English, where there are limited out-of-domain bitext and monolingual data. We approach the low resource problem using two main strategies, leveraging all available data and adapting the system to the target news domain. We explore techniques that leverage bitext and monolingual data from all languages, such as self-supervised model pretraining, multilingual models, data augmentation, and reranking. To better adapt the translation system to the test domain, we explore dataset tagging and fine-tuning on in-domain data. We observe that different techniques provide varied improvements based on the available data of the language pair. Based on the finding, we integrate these techniques into one training pipeline. For \entanosp, we explore an unconstrained setup with additional Tamil bitext and monolingual data and show that further improvement can be obtained. On the test set, our best submitted systems achieve 21.5 and 13.7 BLEU for \taen and \enta respectively, and 27.9 and 13.0 for \iuen and \eniu respectively.

\end{abstract}

\section{Introduction}

We participate in the WMT20 news translation task in two low resource language pairs (four directions), Tamil $\leftrightarrow$ English (\taen and \entanosp) and Inuktitut $\leftrightarrow$ English (\iuen and \eniunosp). These language pairs are challenging due to the lack of in-domain bitext training data and limited monolingual data. For Tamil, the available bitext corpora are from various sources; however, none of the sources is in the news domain, and each corpus is in limited size or noisy. Inuktitut encompasses the challenges present for Tamil, but is even more challenging because the quantity of available monolingual data is even less than the bitext data.

We explore techniques that leverage available data from all languages. First, we investigate supervised learning together with pre-training using mBART~\cite{liu2020multilingual}. Second, inspired by the recent success of improving low resource languages through multilingual models \citep{arivazhagan2019massively,tang2020multilingual}, we explore the utility of multilingual models, in the form of multilingual pretraining and subsequent fine-tuning. Third, we leverage the monolingual data of the source and target languages using data augmentation techniques, such as back-translation~\citep{sennrich2015improving} and self-training~\citep{ueffing2006using, zhang2016exploiting, he2019revisiting}. Following~\citet{chen2019facebook}, we apply these techniques iteratively. Fourth, we use noisy-channel model reranking~\cite{yee-etal-2019-simple} to further boost  performance. The reranking uses language modeling to select a more fluent hypothesis, which requires  monolingual data in the target language.

Additionally, we investigate how adding substantially more unconstrained data can further improve the performance of \enta system. We incorporate data from bitext mining efforts such as \textsc{ccMatrix}~\cite{schwenk2019ccmatrix} and \textsc{ccAligned}~\cite{elkishky2019massive}, as well as additional monolingual data from \textsc{ccNet}~\cite{wenzek2019ccnet} curated from CommonCrawl. The additional data is used for iterative back-translation and to train stronger language models for noisy-channel reranking.

In a complementary direction, we investigate ways to adapt the translation system to the target domain. We explore controlled generation by adding dataset tags to indicate domain. Furthermore, we fine-tune our system on the in-domain data.

For all language directions, we obtain our final systems by fusing a combination of the techniques mentioned above. We observe that the bulk of the improvements in our systems are from iterative back-translation and self-training, except the \eniu system where we only have exceptionally limited quantities of Inuktitut monolingual data. Noisy-channel reranking provides further improvement on top of strong systems, especially for to-English directions where we have high-quality news-domain monolingual data to train a good language model. Each of the other techniques, including dataset tagging, fine-tuning on in-domain data, and ensembling also provides nice improvements. 

\section{Data} \label{sec:data}
For the constrained track, we use monolingual data from all languages provided in WMT20 for mBART pre-training ~\citep{liu2020multilingual}, and we use bitext data between English and other languages for training the system from scratch or fine-tuning the pretrained mBART models.
We also require English, Tamil, and Inuktitut monolingual data for techniques such as back-translation, self-training, and creating language models for noisy-channel reranking. For low resource languages, Tamil and Inuktitut, we use all the available monolingual data, e.g. NewsCrawl + CommonCrawl + Wikipedia dumps for Tamil, and CommonCrawl for Inuktitut. For English, we only use NewsCrawl as the monolingual data because it is sufficiently large, high-quality, and in the news domain.

For the unconstrained track, we use Tamil monolingual data and Tamil-English mined bitext data from external sources based on CommonCrawl. The details are described in Section~\ref{sec:unconstrained_tamil_data}.

\subsection{Data filtering}
\subsubsection{Bitext data} \label{sec:bitext_data}
For each data source for each language pair, we remove duplicate sentence pairs and use \texttt{fastText}~\citep{joulin2016fasttext, joulin2016bag} language identification to remove sentence pairs where either the source or the target sentence is not predicted as the expected language. The resulting size of the bitext data of each language pair is shown in Appendix Table~\ref{tab:bitext_data}.

\subsubsection{Monolingual Data}\label{sec:monolingual_data}
We use monolingual data after \texttt{fastText} language identification filtering from all languages provided in WMT20 to train our mBART model. 
CommonCrawl contains a large quantity of data, but is also quite noisy as it is crawled from the web. Furthermore, the sentences are not in the news domain. 
To clean the data and select the sentences closer to the news domain, we apply the in-domain filtering method described in~\cite{moore-lewis-2010-intelligent} for languages that have NewsCrawl monolingual data. 
First, we train two n-gram language models~\cite{heafield-2011-kenlm} on NewsCrawl and CommonCrawl respectively.
Then, for each sentence from CommonCrawl, we obtain scores from these two language models, compute the difference between normalized log-probability, and we remove the lowest-scoring sentences. We heuristically examine the data and remove the bottom 30\%-60\% of sentences.
Concretely, the scoring function is $ H_{NC}(s) - H_{CC}(s) $, where $s$ is the sentence, $H_{NC}(s)$ and $H_{CC}(s)$ are the word-normalized cross entropy scores for sentence $s$ by n-gram language model trained on NewsCrawl and CommonCrawl data respectively.

We concatenate sentences from different sources and remove duplicate sentences for each language. We show the detailed dataset statistics in Appendix Table~\ref{tab:monolingual_data}.

\subsection{Unconstrained setup for Tamil} \label{sec:unconstrained_tamil_data}

In the unconstrained track, additional data can be used. We incorporate two additional sources of data: noisy bitext from data mining and monolingual data.

\subsubsection{Mined bitext data}

We use mined bitext data from \textsc{ccMatrix}~\cite{schwenk2019ccmatrix} and \textsc{ccAligned}~\cite{elkishky2019massive}, two complementary mining strategies. Both approaches use the web data from unconstrained CommonCrawl to identify noisy bilingual matched pairs. \textsc{ccMatrix} embeds monolingual sentences using \texttt{LASER}~\cite{schwenk2017learning} multilingual sentence embeddings. To identify matching bitext pairs, the distance from each sentence to each other sentence is calculated based on the distance in the embedding space. For \textsc{ccAligned}, documents that could correspond to bitext pairs are aligned first at the document level, then at the paragraph level, and finally at the sentence level. In total, we include 2M aligned English-Tamil mined sentences.

\subsubsection{Monolingual data}


We used additional Tamil monolingual data from CommonCrawl snapshots between 2017-26 to 2020-10 extracted by \textsc{ccNet}~\cite{wenzek2019ccnet}. We break down the document-level structure from \textsc{ccNet} into sentences and apply further processing. We concatenate all the snapshots of the additional monolingual data, deduplicate the sentences, apply \texttt{fastText} language identification and remove sentences are not predicted as Tamil. The final data results in 125M sentences. Subsequently, we concatenate the unconstrained monolingual data with constrained monolingual data, and we use them for back-translation and training Tamil language model.

\section{System overview}
We use the Transformer~\citep{vaswani2017attention} as our model architecture for all of our systems. To better train models with datasets in different sizes, we use random search to select the hyper-parameters that achieve the best BLEU score on the validation set. We use sentencepiece~\citep{kudo2018sentencepiece} to learn the subword units to tokenize the sentences. The details of selected hyper-parameters are listed in Appendix~\ref{sec:hyper_parameters}. All our systems are trained with \texttt{fairseq}\footnote{\url{https://github.com/pytorch/fairseq}}~\citep{ott2019fairseq}.

\subsection{Dataset tag}
Training and decoding the model with an indication of domain (such as a specified dataset tag)~\citep{DBLP:journals/corr/KobusCS16} is a technique that allows us to control the output domain of the trained system. Similarly, \citet{caswell-etal-2019-tagged, chen2019facebook} show that adding specific tag to back-translated and self-translated data can improve model performance. We add dataset tags to all of our systems described in this paper, by pre-pending a domain specific tag to the source sentence during training. At test time, we sweep over all the possible tags that are used during training including ``no tag'', and we choose the tag that achieves the best BLEU score on validation set. We find that when training with dataset tag, the supervised systems are 0.9 and 0.5 BLEU score higher than the system trained without dataset tag for \taen and \enta respectively. See results in Table \ref{tab:dataset_tag}.

\subsection{Baseline systems}
\label{sec:base_systems}
We investigate a variety of baseline approaches as the starting point for our models. For both Tamil and Inuktitut languages, we explore four different baseline systems, (1) bilingual supervised, (2) multilingual supervised, mBART pretraining with (3) bilingual and (4) multilingual fine-tuning. These systems are trained with constrained bitext and monolingual data. We will then improve these baseline models, as described in subsequent sections.

\subsubsection{Bilingual supervised}
To train the base bilingual systems, we pre-pend the dataset tag to the source sentence to differentiate data from different corpus and concatenate all data sources for that language. 

\subsubsection{Multilingual supervised}
\citet{arivazhagan2019massively} shows that multilingual model can improve the model performance of medium and low resource languages, as multilingual models are often trained on greater quantities of data compared to bitext models. Thus, we investigate if multilingual supervised models can be stronger starting points. We use all the bitext data between English and other languages provided in WMT20 to train many-to-one (XX $\rightarrow$ English) and one-to-many (English $\rightarrow$ XX) models. One challenge of multilingual training is different language directions have different quantities of data, and the high resource language can starve for capacity while low resource language can benefit from the transfer. To balance the trade-off between learning and transfer,
we follow~\citet{arivazhagan2019massively} with a temperature-based strategy to sample sentences from different languages. Furthermore, for each direction, we optimize the transfer by selecting the best temperature and model checkpoint based on the BLEU score of the target language pair validation set. 

\subsubsection{mBART-pretraining with bilingual and multilingual fine-tuning} \label{sec:mbart_fine_tuning}
For mid and low resource languages, the quantity of available bitext may be low, but large resources of monolingual data exist. This monolingual data can be used in the form of pre-training, followed by subsequent fine-tuning into translation models. We use mBART~\cite{liu2020multilingual} -- a multilingual denoising pre-training approach -- to pre-train our systems, which has shown substantial improvements compared to training the model from scratch.
First, we pre-train mBART across $13$ languages (Cs, De, En, Fr, Hi, Iu, Ja, Km, Pl, Ps, Ru, Ta, Zh) on all monolingual data provided by WMT 20. For pretraining, we used a batch size of $2048$ sequences per batch and trained the model for $240K$ steps. We learn the SPM jointly on all languages. We sample the same amount of sentences from monolingual data of all languages to learn a vocabulary of $130,000$ subwords.
In the fine-tuning stage, we use exactly the same data sources as the bilingual supervised model and multilingual supervised model. For multilingual fine-tuning, previously people have built bitext translation systems by fine-tuning pretrained mBART models. Recent work~\citet{tang2020multilingual} extended this to multilingual fine-tuning, which can create multilingual translation models from multilingual pre-trained models. Different from \citet{tang2020multilingual}, we tune the temperature rate separately for the four language directions we focus on. In the multilingual fine-tuning stage, we use random search to sweep over dropout, learning rate, and temperature sampling factor, and we select the model checkpoint based on the BLEU score evaluated on the target language pair validation set. 

\subsection{Iterative back-translation (BT)}
Back-translation~\citep{sennrich2015improving} is an effective data augmentation technique to improve model performance with target side monolingual data. The method starts from training a target to source translation system, which is subsequently used to translate the monolingual data in the target language back to source language. Then the synthetic back-translated dataset is concatenated with the raw bitext data to train the source to target translation model. After the source to target model is improved, the same technique can be applied again to train the back-translation system in the reversed direction. We repeat the process for several iterations until no significant improvement is obtained.

In all of our back-translation systems, we follow ~\citet{chen2019facebook} to add dataset tags to both raw bitext data and back-translated data. We upsample the bitext data, and the upsampling ratio is selected based on parameter sweeping and validating the resulting improvement on the validation set. Beam search with beam size 5 is used when generating the synthetic sentences.

\subsection{Noisy-channel reranking (NCD)} \label{sec:ncd}
Reranking is a technique that uses a separate model to score and better select hypotheses from the n-best list generated by the the source to target model. To rerank our system output, we use the noisy-channel model ~\citep{yee-etal-2019-simple} as the scoring model~\citep{ng-etal-2019-facebook, chen2019facebook}. Given a source and target sentence pair (x, y), the noisy-channel model scores it with
\begin{equation}
\log P(y|x) + \lambda_1 \log P(x|y) + \lambda_2 \log P(y) \label{eq:nc}
\end{equation}
where $\log P(y|x)$, $\log P(x|y)$ and $\log P(y)$ are the forward model, backward model and language model scores. The weights, $\lambda_1$ and $\lambda_2$, are tuned through random search on the validation set. All of our submitted test set hypotheses are ranked and selected by noisy-channel reranking.

The language models used in noisy-channel reranking are Transformers. For constrained track, we use the monolingual data as described in Section~\ref{sec:data} to train the language models for English, Tamil. For Inuktitut, we find that the monolingual data is very limited and even smaller than the size of bitext data, therefore we concatenate the CommonCrawl data with the Inuktitut side of the bitext data together to train the Inuktitut language model.
For unconstrained Tamil language model, we train on the constrained data with the additional unconstrained data extracted by \textsc{ccNET} as described in Section~\ref{sec:unconstrained_tamil_data}. The SPM size, model hyper-parameters, and evaluation of the language models can be found in Appendix~\ref{appendix:lm}.

\subsection{Self-training (ST)}
Self-training~\citep{ueffing2006using, zhang2016exploiting, he2019revisiting} is a method that leverage monolingual data in source language to improve the system performance. We use the trained source to target translation system to translate monolingual data in source language to target language. Similar to BT, the synthetic dataset can be concatenated with bitext data to train the source to target model again. We follow ~\citet{chen2019facebook} and use the noisy-channel model to select the top synthetic sentence when decoding from monolingual data into the source language. We inject the same types of noise to the source side of synthetic data as~\citet{he2019revisiting}. 

\citet{shen2019sourcetarget,chen2019facebook} both show that self-training can provide complementary improvement in addition to back-translation, especially when (1) there is lack of target side monolingual data, (2) source side monolingual data is much similar to the domain of test set compared with target side monolingual data, and (3) the decoding method outperforms greedy decoding on the source to target model. Therefore, we experiment self-training on \eniu due to greater quantities of in-domain source side monolingual data, on \iuen in Nunavut Hansard domain with Inuktitut side of bitext data due to much more in-domain monolingual data on the source side, and on \taen because we observe great improvement from noisy-channel reranking. However, we only observe significant improvement on \taen system.

\subsection{Fine-tuning (FT) on validation set}

Fine-tuning is a technique to adapt the model to the target domain when the initial model is not trained with training data in the target domain. In both Tamil and Inuktitut, none of the training data is in news domain as the test data, therefore we fine-tune our final systems on a portion of the validation data and evaluate on the rest of hold-out validation data.
For Tamil systems, we split the validation data with a 75-25 split, where 75\% of the data is used for fine-tuning and 25\% of the data is used for evaluation. \taen and \enta systems are fine-tuned and evaluated on the same split of validation dataset.
For Inuktitut systems, we split the validation set based on the domain --- Nunavut Hansard or news. For each domain, we split the validation data with a 75-25 split for fine-tuning and evaluation. We fine-tune our best performing \iuen and \eniu systems in domain on the corresponding validation set split.

\begin{table}[t]
\centering
\small
\begin{tabular}{l|cccc}
\toprule
Model & \taen & \enta & \iuen & \eniu \\
\midrule
w/o tag & 15.6 & 8.5 & 31.4 & 16.1 \\
with tag & 16.5 & 9.0 & 31.3 & 16.1 \\
\bottomrule
\end{tabular}
\caption{\small Systems trained with and w/o dataset tags. The BLEU score is reported on validation set. We sweep all available dataset tags when decoding on validation set and report the best performing dataset tag. The BLEU scores of each dataset tag are reported in Appendix \ref{appendix:dataset_tag_decoding}}
\label{tab:dataset_tag}
\end{table}

\section{Results}
In this section, we describe the details of our systems, and we report \textsc{SacreBLEU} ~\citep{post-2018-call} on the validation set for intermediate iterations and ablations. For our validation set fine-tuned systems, we report the BLEU score on our validation holdout set split. Our general strategy for all language directions was to identify the best performing baseline setting, then iteratively improve upon the baseline using back-translation and self-training. Finally, we apply noisy-channel reranking and fine-tuning on validation set to create our final submission. 

\subsection{Baseline} \label{sec:baseline}
We explore four different baseline approaches as described in Section~\ref{sec:base_systems} for each language direction in the constrained setup, Inuktitut $\leftrightarrow$ English and Tamil $\leftrightarrow$ English. The detailed results are shown in Table \ref{tab:baseline_systems}.

First, bilingual models are trained with bilingual bitext data.
Next, we focus on multilingual training. The multilingual supervised models are trained with all the available bitext data provided by WMT20. We use the same SPM as described in Section~\ref{sec:mbart_fine_tuning}.
For both bilingual and multilingual models, we initialize the model weights either randomly or with pre-trained mBART model weights. Therefore, for each language direction, we have four combinations, bilingual supervised, multilingual supervised, mBART + bilingual fine-tuning and mBART + multilingual fine-tuning. We use dataset tags for all systems, and we sweep the tag that performs the best when decoding on the validation set. Additional details and hyper-parameters are provided in the Appendix~\ref{sec:hyper_parameters}.

For to-English directions, both multilingual models and mBART pretraining can get better model performance than bilingual supervised model as shown in Table~\ref{tab:baseline_systems}. For \taen direction, mBART + multilingual fine-tuning performs the best with 20.4 BLEU, which outperforms bilingual supervised system by 3.2 BLEU score. For the \iuen direction, mBART + bilingual fine-tuning works the best and gets 32.9 BLEU score, which outperforms bilingual supervised baseline by 2.8 BLEU score. However, for from-English directions, we do not observe similar advantages with either multilingual model or mBART pretraining, and a properly tuned bilingual supervised model achieves the best results for both directions. We get 8.0 BLEU score for \enta direction, and we get 16.1 BLEU score for \eniu direction.

\begin{table}[t]
\centering
\small
\begin{tabular}{l|cccc}
\toprule
System & \taen & \enta & \iuen & \eniu \\
\midrule
bi.  & 17.2 & 8.0 & 29.7 & 16.1 \\
multi. & 18.2 & 7.1 & 30.7 & 15.8 \\
bi-FT$^\star$ & 18.9 & 8.0 & 32.9 & 16.1 \\
multi-FT$^\star$ & 20.4 & 7.4 & 32.5 & 16.0 \\
\bottomrule
\end{tabular}
\caption{\small BLEU scores of baseline systems evaluated on the validation set. $^\star$ Pre-trained on mBART.}
\label{tab:baseline_systems}
\end{table}

\subsection{Tamil systems}
\subsubsection{Constrained \taen system} \label{sec:constrained_ta_en}

For the \taen system, we first use the \enta bilingual baseline system (ensemble) to generate back-translation data from English NewsCrawl data. We then train our first iteration back-translation system (``iter1-BT") with upsampled bitext (upsampling ratio tuned on the validation set). Similarly, we train our second iteration back-translation system (``iter2-BT") with upsampled bitext and back-translation data generated by \enta iter1-BT system (ensemble). The iter2-BT system (ensemble) is then used to generate ST data from Tamil NewsCrawl, CommonCrawl and Wiki data. We combine it with iter2-BT system's data to train the iter2-BT+ST system. Finally, we fine-tune this system on the validation set and apply noisy-channel reranking to select the hypotheses. We explore Transformer models of different capacities and choose Transformer \emph{big} (with 8K feed-forward dimension) for a good balance of performance and training speed. For the iter2-BT+ST system (and its ensemble/finetuned version), we further enlarge the encoder to 10 layers given higher data abundance. We can see from Table~\ref{tab:tamil_systems} that our training pipeline improves model performance steadily ($\geq1.3$ validation BLEU) after iterations, and in-domain fine-tuning as well as noisy-channel reranking are very helpful to alleviate the effects of train-test domain mismatch.

\subsubsection{Constrained \enta system} \label{sec:constrained_en_ta}
For the \enta system, we first use the mBART+multi-FT baseline system for \taen to generate back-translation data from the monolingual data. We add different back-translation dataset tags based on the source of monolingual data and train our first iteration back-translation system (``iter1-BT") by tuning upsampling ratios on the bitext and back-translation datasets. For the model architecture, we explore the options of training Transformers from scratch and fine-tuning a pretrained mBART model and find that the former performs better with ensembles. Doing one iteration of training with back-translation data gives 5.8 BLEU increase (Table~\ref{tab:tamil_systems}). We further train the second iteration back-translation system (``iter2-BT") with back-translation data generated from the best iter1-BT \taen system. As the gain from the second iteration is small (0.4 BLEU), we do not continue for the third iteration. Noisy-channel reranking is applied with the best systems from both language directions and the Tamil language model (Appendix \ref{appendix:lm}). We observe little gain (0.1 BLEU) and suspect it’s due to the high perplexity of the language model. Further fine-tuning the iter2-BT model on the validation set gives 4.1 BLEU score improvement on the validation holdout set.

\begin{table}[t]
\centering
\small
\begin{tabular}{l|cc}
\toprule
System & \taen & \enta \\
\midrule
baseline & 20.4 & 8.0  \\
 + ensemble & 21.2 & 9.0 \\
iter1-BT & 23.4 & 13.8 \\
 + ensemble & 24.8 & 14.1 \\
iter2-BT & 25.6 & 14.2 \\
 + ensemble & 26.4 & 14.3 \\
 + NCD & 28.5 & 14.4 \\
\midrule
\multicolumn{3}{l}{eval on valid holdout} \\
\midrule
iter2-BT     & 26.2 & 14.6 \\
iter2-BT+ST  & 27.5 & - \\
iter2+FT on valid & 28.0 & 18.7 \\
 + ensemble & 28.3 & 19.0 \\
 + NCD & 29.8 & 19.5 \\
\midrule
\multicolumn{3}{l}{unconst. eval on valid holdout} \\
\midrule
iter2-BT & - & 15.2 \\
iter2-BT+FT & - & 19.6 \\
  + ensemble & - & 19.6 \\
  + NCD & - & 20.2 \\
\bottomrule
\end{tabular}
\caption{\small Results of Tamil systems. We report the BLEU scores on newsdev2020 validation set.}
\label{tab:tamil_systems}
\end{table}

\begin{table}[h]
\centering
\small

\begin{tabular}{l|c|c|c}
\toprule
\multicolumn{1}{c|}{system} & \multicolumn{3}{c}{\iuen}\\
\midrule
                           & NH   & News  & Combined \\
\midrule
baseline                   & 42.4 &    19.2  &   32.9   \\
+ ensemble                 & 42.4 &    19.4  &   32.9    \\
iter1-BT                   & 43.3 &    24.1  &   35.1    \\
+ ensemble                 & 43.8 &    24.6  &   35.7    \\
\hline
\multicolumn{3}{l}{eval on valid holdout} \\
\hline
iter1-BT                   & 46.1 & 24.3 & 35.0 \\
iter1-BT+FT on valid     & 47.3 & 31.1 & 38.4 \\
+ ensemble                 & 48.2 & 31.7 & 39.2 \\
+ NCD                      & 49.0 & 32.8 & 40.2 \\
\bottomrule
\end{tabular}
\caption{\small Results of \iuen systems. We report BLEU scores on both domain-split and the whole newsdev2020 validation set}
\label{tab:iuen_systems} 
\end{table}

\begin{table}[h]
\centering
\small

\begin{tabular}{l|c|c|c}
\toprule
\multicolumn{1}{c|}{system} & \multicolumn{3}{c}{\eniu} \\
\midrule
                           & NH   & News  & Combined  \\
\midrule
baseline                   & 24.5   &    5.3  &   16.1    \\
+ ensemble                 & 24.8   &    5.6  &   16.3    \\
iter1                      & 24.8 (ST)   &    5.5 (BT)  &   16.3    \\
+ ensemble                 & 25.0 (ST)   &    5.8 (BT)  &   16.5    \\
\hline
\multicolumn{3}{l}{eval on valid holdout} \\
\hline
iter1                   & 27.6 (ST) & 5.4 (BT) & 15.5 \\
iter1+FT on valid       & 28.9 & 14.5 & 20.8  \\
+ ensemble              & 28.9 & 15.1 & 21.1 \\
+ NCD                   & 28.9 & 16.6 & 22.0 \\
\bottomrule
\end{tabular}
\caption{\small Results of \eniu systems. We report BLEU scores on both the domain-split and whole newsdev2020 validation set.}
\label{tab:eniu_systems} 
\end{table}

\subsubsection{Unconstrained \enta system} 

For the unconstrained track, we first used the iteration1 + back-translation ensemble model to back-translate the additional monolingual data from CommonCrawl. Subsequently, we combined back-translated data from unconstrained monolingual sources with back-translated data from WMT monolingual data from English and Tamil, with the WMT bitext and mined \taen data. We used the same BPE and vocabulary as the constrained system. The data was deduplicated, and the validation and test data removed if an exact match was present in the training data. The mined data was additionally cleaned to remove sentences longer than 250 BPE tokens, as well as bitext pairs where the length between the source and target was greater than 2.5x difference. Subsequently, we trained a large Transformer sequence-to-sequence model on the total combined data using various data domain tags. After training was complete, we further fine-tuned on the validation set, as described in Section~\ref{sec:constrained_en_ta}. We applied noisy-channel reranking when decoding test data. The forward model is ensembled with two of the best performing fine-tuned models. The backward model is the best performing model in Section \ref{sec:constrained_ta_en}, which is ensembled with two fine-tuned models. The language model is unconstrained Tamil language model described in Section~\ref{sec:ncd}. We rerank from best 20 hypothesises generated by ensembled forward model, and we achieve 20.2 BLEU score on validation set.

\subsection{Inuktitut systems}
The Inuktitut validation and test set are composed of data from two different domains, the proceeding of the Legislative Assembly of Nunavut from Nunavut Hansard (NH) and news. We find that the model can be further improved if we optimize our translation training pipeline for these two domains separately, and therefore we train and report BLEU score separately for each domain. We also report the BLEU score on the whole validation set, where we use the domain-specific system to decode on the portion of the corresponding domain, concatenate the hypothesises and compute the BLEU score.

\subsubsection{Constrained \iuen systems} \label{sec:constrained_iuen}
For the \iuen~system, we use \eniu bilingual supervised system described in Section \ref{sec:baseline} for back-translation. The model used for decoding is an ensemble of 3 \eniu models, and we decode from the English NewsCrawl data. We concatenate the back-translated data with bitext data and sweep the upsampling ratio of the bitext data to find the best ratio. We experiment with both mBART pretraining + bilingual fine-tuning and training from scratch, and we find that mBART + bilingual fine-tuning works better on Nunavut Hansard domain of validation set, and training from scratch works better on news domain. The hypothesis is that the English NewsCrawl monolingual data for back-translation is in-domain with the news domain validation set and there is huge amount of English NewsCrawl data, so the advantage of pretraining is not significant. We also experiment with self-training on \iuen direction in Nunavut Hansard domain, where we use the source to target model (ensembled) to decode from the Inuktutit side of Nunavut Hansard 3.0 parallel corpus with noisy-channel reranking; however, we do not observe any improvement. The best result at the first iteration is from the back-translation system, which outperforms baseline system by 2.2 BLEU score (Table~\ref{tab:iuen_systems}), where most of the gain comes from improvement on news domain. 

We do not observe gains for doing the second iteration of back-translation for \iuen system, and we suspect that it is due to lack of improvement for our \eniu model from supervised approach to the first iteration. We then fine-tune the best iteration 1 \iuen models on validation data for each domain. The final domain-specific systems are ensembled from the fine-tuned models and followed by noisy-channel reranking. To use noisy-channel reranking for Nunavut Hansard domain, we fine-tune the English language model described in~\ref{sec:ncd} on English side of the Nunavut Hansard 3.0 training data provided in WMT20. The best \iuen system we submit has 40.2 BLEU score on our validation holdout set.

\subsubsection{Constrained \eniu systems}
We experiment with both self-training and back-translation with the best baseline systems reported in \ref{sec:baseline} to improve \eniu system. For self-training, we use ensembled supervised \eniu model and beam decoding with beam size 5 to decode from English monolingual data. We decode from the English side of Nunavut Hansard 3.0 parallel corpus to train the model for Nunavut Hansard domain, and we decode from the English NewsCrawl data for news domain. However, we do not observe improvement for news domain, and there is only mild improvement (0.3 BLEU) for Nunavut Hansard domain as shown in Table~\ref{tab:eniu_systems}.
For back-translation, we use iteration 1 \iuen news domain model from \ref{sec:constrained_iuen} to decode constrained Inuktitut CommonCrawl data. We get no improvement on Nunavut Hansard domain and mild improvement (0.2 BLEU) on news domain. We use self-training system for Nunavut Hansard domain and back-translation system for news domain, and it achieves 16.3 BLEU score on the validation set, which is merely 0.2 BLEU score improvement over baseline system. 
We then fine-tune the best systems we get on domain-specific validation set splits, followed by ensembling and noisy-channel reranking. The fine-tuning is very effective for the news domain, where we get 9.1 BLEU score improvement. This is expected because we do not have any training data from news domain. Our final submitted system achieves 22.0 on our validation holdout set.

\begin{table}[h]
\centering
\small
\begin{tabular}{l|c}
\toprule
  Submitted system & BLEU  \\
\midrule
  \taen & 21.5 \\
  \enta & 12.6 \\
  \enta (unconst.) & 13.7 \\
  \iuen & 27.9 \\
  \eniu & 13.0 \\
\bottomrule
\end{tabular}
\caption{\small Results of our best submitted systems of each direction. We report BLEU scores on newstest2020.}
\label{tab:final_submission} 
\end{table}

\section{Conclusion}

This paper describes Facebook AI’s Transformer based translation systems for the WMT20 news translation shared task. We focused on two low-resource languages pairs, Tamil $\leftrightarrow$ English and Inuktitut $\leftrightarrow$ English, and we explored the same set of techniques, including dataset tagging, mBART pretraining and fine-tuning, back-translation and self-training, fine-tuning on domain-specific data, ensembling, and noisy-channel reranking.
We demonstrated strong improvements by stacking these techniques properly on three language directions, \taen, \enta, and \iuen. The \eniu direction is difficult to improve due to lack of target side monolingual data. Surprisingly, self-training does not work on \eniu either even we have huge amounts of in-domain English side monolingual data. We are interested in continued exploration on how to better leverage source side monolingual data to improve \eniu and other low resource languages where we do not have enough target side monolingual data.

\section{Acknowledgements}
We thank Marc'Aurelio Ranzato for providing discussion and guidance during the competition, Vishrav Chaudhary for sharing insightful data cleaning approaches, Guillaume Wenzek for previous work on ccNET for monolingual data used in unconstrained setting, Yuqing Tang for the work of mBART pretraining and multilingual fine tuning, Ahmed El-Kishky and Holger Schwenk for sharing their mined data for Tamil, Sergey Edunov for sharing cleaned up dataset to speed up our early exploration, and Michael Auli for sharing experience about noisy-channel reranking technique.

\bibliographystyle{acl_natbib}
\bibliography{anthology,emnlp2020}

\appendix

\newpage
\clearpage

\section{Constrained data} \label{appendix:data}
In this section, we list the statistics for all the constrained datasets we use to build for our systems.
\paragraph{Bitext data}
Table~\ref{tab:bitext_data} shows the bitext data we used for multilingual systems. We use all bitext data between English and other 11 languages provided in WMT 20 except a couple of sources. We do not include the data back-translated by other system to avoid introducing bias. We do not include CzEng 2.0 for Czech nor CCMT for Chinese due to human mistake. We follow the filtering steps described in Section~\ref{sec:bitext_data}, and the size of dataset for each language pairs are listed in Table~\ref{tab:bitext_data}.
\paragraph{Monolingual data}
Table~\ref{tab:monolingual_data} shows the list of monolingal data we use for mBART-pretraining with 13 languages. We follow Section~\ref{sec:monolingual_data} to filter the monolingual data, and we list the amount of data before and after the filtering step.

\section{Language model used in noisy-channel reranking} \label{appendix:lm}
Language model is required in the noisy-channel reranking system. We learn the BPE subwords with sentencepiece, and we train the Transformer based causal language models with \texttt{fairseq} in fp16 mode. The model size and hyper-parameters are tuned based on the perplexity of newsdev2020 validation sets per language. We describe the data and hyper-parameters of each language below, and we report the perplexities in Table~\ref{tab:language_models}.

\paragraph{English language model}
We train our English language model with the high quality NewsCrawl data provided by WMT 20. We use the same filtering steps in Section~\ref{sec:monolingual_data} for NewsCrawl. We learn the BPE with 32K vocabulary size.
We train the transformer-based model with 36 transformer layers, 1280 embedding dimension size, 5120 ffn dimension size, 20 attention heads and resulting in 749M parameters. The optimizer is Adam~\citep{kingma:adam:2015} optimizer with $beta1 = 0.9$ and $beta2 = 0.98$. We use polynomial decay learning rate scheduler with 0.005 learning rate and 0.1 dropout rate. The maximum tokens are 4096 for each batch per GPU, and we train with 64 GPUs for 58K updates. As we show in Table~\ref{tab:language_models}, this model achieves 23.3 perplexity on English side of Ta-En newsdev2020 set, 25.3 perplexity on news portion of Iu-En newsdev2020 set, and 29.7 perplexity on Nunavut Hansard portion of Iu-En newsdev2020 set. The perplexity on news validation sets are lower than none-news validation set. We use the English language model to rerank \taen system and news domain of \iuen system.

To better rerank \iuen hypothesises for Nunavut Hansard domain, we fine-tune the English language model on English side of Nunavut Hansard 3.0 parallel corpus. The perplexity on Nunavut Hansard portion of Iu-En newsdev2020 set is significantly improved from 29.7 to 8.1. We use the fine-tuned English language model to rerank the Nunavut Hansard domain of \iuen system.

\paragraph{Tamil language model}
We train the Tamil language model for constrained \enta system with all the available Tamil monolingual data preprocessed in Section~\ref{sec:monolingual_data}. The BPE vocabulary size is 32K. We train the transformer-based language model with 24 transformer layers, 1024 embedding size, 4096 ffn embedding size, 16 attention heads and resulting in 335M parameters. We use Adam optimizer with $beta1 = 0.9$ and $beta2 = 0.98$. We use polynomial decay learning rate scheduler with 0.005 learning rate and 0.1 dropout rate. The maximum tokens are 8192 for each batch per GPU, and we train with 16 GPUs for 46K updates. The model achieves 61.8 perplexity on Tamil side of Ta-En newsdev2020 set.

For unconstrained \enta system, we use both constrained Tamil monolingual data and the additional Tamil monolingual data described in Section~\ref{sec:unconstrained_tamil_data}. We share the same 32K BPE vocabulary as constrained Tamil language model. We use a larger transformer model with 32 transformer layers, 1024 embedding size, 4096 ffn embedding size, 8 attention heads. We use Adam optimizer with $beta1 = 0.9$ and $beta2 = 0.98$. We use cosine learning rate scheduler with 0.0001 learning rate and 0.3 dropout rate. The maximum tokens are 3072 for each batch per GPU, and we train with 32 GPUs for 69K updates. The model achieves 40.6 perplexity on Tamil side of Ta-En newsdev2020 set, which is better than the constrained Tamil language model.

\paragraph{Inuktitut langauge model}
The Inuktitut language model is trained with Inuktitut side of Nunavut Hansard 3.0 parallel corpus and the constrained Inuktitut monolingual data provided by WMT 20. The BPE vocabulary size is 5K. We train the transformer-based language model with 6 transformer layers, 512 embedding size, 4096 ffn embedding size, 8 attention heads and resulting in 34M parameters. We use Adam optimizer with $beta1 = 0.9$ and $beta2 = 0.98$. We use inverse square root learning rate scheduler with 0.0005 learning rate and 0.3 dropout rate. The maximum tokens 2048 for each batch per GPU, and we train with 8 GPUs for 89K updates. The model achieves 34.9 perplexity on Nunavut Hansard domain of Iu-En newsdev2020 set, and 81.69 perplexity on news portion of Iu-En newsdev2020 set.

\section{The effect of dataset tag at decoding time} \label{appendix:dataset_tag_decoding}
We train our systems with dataset tag, and we sweep the dataset tags by add different tags to the same validation set and select the best performing tag. Table~\ref{tab:dataset_tag_tamil} and \ref{tab:dataset_tag_inuktitut} show the system performance across different dataset tags.

First, we observe that sweeping the best performing dataset tag at decoding time is necessary. Using ``no tag'' to decode works the best for both \taen and \enta systems; however, using specific dataset tags works better for \iuen and \eniu systems. Second, the large BLEU score variations when decoding with different dataset tags show that the tags help the model to better adapt to different domains.

Overall, systems trained with dataset tags works better than trained without dataset tag as we show in Table~\ref{tab:dataset_tag}.

\section{Hyper-Parameters} \label{sec:hyper_parameters}
In this section, we report the hyper-parameters we use. For all of our translation systems, we use transformer based encoder-decoder model with shared embedding across encoder, decoder input and output embedding. We use Adam optimizer with $beta1 = 0.9$ and $beta2 = 0.98$, inversed square root learning rate scheduler, and 4000 warm-up steps with linearly increased rate. The loss is cross-entropy with label smoothing~\citep{labelsmoothing}. We use the same batch sizes with maximum number of tokens 4096, and all models are trained with fp16. We sweep other hyper-parameters with random search, and we select the best performing system based on the evaluated BLEU scores on validation sets.

\paragraph{mBART pretraining}
We train the denoising mBART model with the constrained monolingual data from 13 languages described Section~\ref{sec:monolingual_data}. We learn joint BPE across all languages with vocabulary size 130K. The transformer based encoder-decoder model has 12 encoder and decoder layers, 1024 embedding dimension, 4096 ffn embedding dimension and 16 attention heads, resulting in 487M parameters. We train the model with 0.0003 learning rate, 0.1 dropout rate, and no label-smoothing. We train the model with 256 GPUs for 240K updates.

\paragraph{Tamil systems} For \taennosp, the best performing systems are mBART+multilingual fine-tuning model for baseline system, back-translation system for iteration 1 and BT+ST system for iteration 2. We report the hyper-parameters of the best performing system at each iteration in Table~\ref{tab:hyperparameters_taen}.

For \entanosp, the best performing systems are bilingual supervised model for baseline system, back-translation system for iteration 1 and iteration 2. We report the hyper-parameters of the best performing system at each iteration, including the unconstrained system in Table~\ref{tab:hyperparameters_enta}.

\paragraph{Inuktitut systems}
For \iuennosp, the best baseline system is the mBART pretraining with bilingual fine-tuning. In iteration 1, we tune the model separately for Nunavut Hansard domain and news domain. The best Nunavut Hansard domain model is mBART pretraining with bilingual fine-tuning on bitext and news back-translated data, and the best news domain model is the back-translation model train from scratch.
For \eniunosp, the best baseline system is bilingual supervised model. Similar to \iuen system, we tune the model separately for Nunavut Hansard domain and news domain in iteration 1. The best system for Nunavut Hansard domain is self-training model train from scratch, and the best system for news domain is the back-translation model train from scratch. We report the hyper-parameters of the best performing \iuen and \eniu systems at each iteration in Table~\ref{tab:hyperparameters_iuen} and \ref{tab:hyperparameters_eniu}.

\clearpage
\newpage


\setcounter{table}{0}
\renewcommand{\thetable}{A.\arabic{table}}

\begin{table*}[t]
\centering
\small
\begin{tabular}{|c|c|c|l|}
\hline
\multicolumn{1}{|l|}{\multirow{2}{*}{Language pair}} & \multicolumn{2}{c|}{\# of sentences (M)} & \multicolumn{1}{l|}{\multirow{2}{*}{Missing datasets}} \\
    \cline{2-3}
                           & Raw  & Cleaned                                    &    \\
\hline
Cs-En                   &  9.3   & 8.6    & CzEng2.0, back-translated news    \\ \hline
De-En                   &  48   &   45.9    &           \\ \hline
Hi-En                   &  1.48   &   1.27    &   \\ \hline
Iu-En                   &  0.77   &   0.77    &  \\ \hline
Ja-En                   &  18.2   &   16.2   &  \\ \hline
Km-En                   &  4.4   &   2.46    &  \\ \hline
Pl-En                   &  11.6   &   10.6    &  \\ \hline
Ps-En                   &  1.13   &   0.58    &  \\ \hline 
Ru-En                   &  43.5   &   32.8    & back-translated news  \\ \hline
Ta-En                   &  0.71   &   0.62    &  \\ \hline
Zh-En                   &  19.6   &   15.8    & CCMT, back-translated news \\ \hline
\end{tabular}
\caption{\small En-XX bitext data used for bilingual and multilingual systems. For each language pair, we use all available sources released in WMT20 except the datasets that are listed in the table.}
\label{tab:bitext_data}
\end{table*}

\begin{table*}[h]
\centering
\small
\begin{tabular}{|c|c|c|l|}
\hline
\multicolumn{1}{|l|}{\multirow{2}{*}{Language}} & \multicolumn{2}{c|}{\# of sentences (M)} & \multicolumn{1}{l|}{\multirow{2}{*}{Sources}} \\
    \cline{2-3}
                           & Raw  & Cleaned                                    &    \\
\hline
Cs                   &  355    &  287     & NCL, NC, CC          \\ \hline
De                   &  3528   &  1355    & NCL, NC, EP, CC          \\ \hline
En                   &  4264   &  2685     & NCL, NC, ND, EP, CC, Wiki  \\ \hline
Fr                   &  5853   &   1455    & NCL, NC, ND, EP, CC  \\ \hline
Hi                   &  45     &   43.4    & IITB  \\ \hline
Iu                   &  0.9   &   0.9    & \makecell[l]{Nunavut Hansard parallel corpus 3.0, \\ CC } \\ \hline
Ja                   &  1776 &   1182    & NCL, NC, CC \\ \hline
Km                   &  12.1 &   11.3    & CC, Wiki \\ \hline
Pl                   &  1459 &   1183    & NCL, EP, CC \\ \hline
Ps                   &  5.9 &   5.4    & CC, Wiki \\ \hline
Ru                   &  1261 &   665    & NCL, NC, CC  \\ \hline
Ta                   &  30.3 &   29.4    & NCL, CC, Wiki \\ \hline
Zh                   &  1677 &   806    & NCL, NC, CC \\ \hline
\end{tabular}
\caption{\small Monolingual data used for mBART pretraining and back-translation. The abbreviation in the sources column represent the following, CC: CommonCrawl, EP: Europarl, NC: NewsCommentary, NCL: NewsCrawl, ND: NewsDiscussions, Wiki: Wikipedia}
\label{tab:monolingual_data}
\end{table*}

\setcounter{table}{0}
\renewcommand{\thetable}{B.\arabic{table}}
\begin{table*}[h]
\centering
\scalebox{0.75}{
\begin{tabular}{|l|l|l|l|l|l|l|}
\hline
\multirow{2}{*}{Target language} & \multicolumn{2}{l|}{Training data} & \multirow{2}{*}{BPE size} & \multicolumn{3}{l|}{PPL on newsdev2020} \\ \cline{2-3} \cline{5-7} 
 & source & \# of sentences &  & Ta-En & Iu-En (NH) & Iu-En (news) \\ \hline
English & NewsCrawl & 190M & \multirow{2}{*}{32K} & 23.3 & 29.7 & 25.3 \\ \cline{1-3} \cline{5-7} 
\makecell[l]{+ FT on \\ ~~English side of NH} &  &  & & 77.6 & 8.1 & 27.1 \\ \hline \hline
Tamil & \makecell[l]{CommonCrawl, NewsCrawl, \\ Wikipedia} & 30M & \multirow{2}{*}{32K} & 61.8 & - & - \\ \cline{1-3} \cline{5-7} 
unconst. Tamil & \makecell[l]{constrained Tamil data, \\ CommonCrawl in Sec.~\ref{sec:unconstrained_tamil_data}} & 155M & & 40.6 & - & - \\ \hline \hline
Inuktitut & \makecell[l]{Inuktitut side of Nunavut Hansard 3.0, \\ CommonCrawl} & 860K & 5K & - & 34.9 & 81.7 \\ \hline
\end{tabular}
}
\caption{\small Statistics of language models for each language. }
\label{tab:language_models}
\end{table*}

\setcounter{table}{0}
\renewcommand{\thetable}{C.\arabic{table}}
\begin{table*}[!htbp]
\centering
\small
\begin{tabular}{l|cc}
\toprule
Tag & \taen & \enta \\
\midrule
None & 16.5 & 9.0 \\
mkp  & 15.4 & 8.0 \\
nlpc & 15.6 & 6.8 \\
pib & 15.5 & 8.6 \\
pmindia & 15.5 & 8.7 \\
tanzil & 11.9 & 0.6 \\
ufal & 16.1 & 8.2 \\
wikimatrix & 4.0 & 6.4 \\
wikititles & 15.8 & 8.5 \\

\bottomrule
\end{tabular}
\caption{\small Tamil bilingual supervised single model performance when decoding on validation set with different dataset tags. The BLEU score is evaluated newsdev2020 validation set. }
\label{tab:dataset_tag_tamil}
\end{table*}

\begin{table*}[!htbp]
\centering
\small
\begin{tabular}{l|cc}
\toprule
Tag & \iuen & \eniu \\
\midrule
None & 29.7 & 15.8 \\
Nunavut Hansard & 31.3 & 16.0 \\
wikititles & 30.1 & 16.1 \\

\bottomrule
\end{tabular}
\caption{\small Inuktitut bilingual supervised single model performance when decoding on validation set with different dataset tags. The BLEU score is evaluated on newsdev2020 validation set.}
\label{tab:dataset_tag_inuktitut}
\end{table*}

\setcounter{table}{0}
\renewcommand{\thetable}{D.\arabic{table}}

\begin{table*}[t]
\small
\centering
\scalebox{0.85}{
\begin{tabular}{|l|l|l|l|l|l|l|l|l|l|l|}
\hline
System & \makecell[c]{Subword \\ (size)} & \# params & layers & \makecell[c]{embed \\ size} & \makecell[c]{ffn \\ embed size} & \makecell[c]{attention \\ heads} & \makecell[c]{learning \\ rate} & dropout & \makecell[c]{label \\ smoothing} & \# GPUs \\ \hline
\makecell[l]{Baseline system \\ (mBART+multi-FT)} & \makecell[c]{BPE \\ (130K)} & 487M & 12 & 1024 & 4096 & 16 & 0.0001 & 0.2 & 0.2 & 16 \\ \hline
iter1 (BT) & \makecell[l]{Unigram \\ (16384)} & 293M & 6 & 1024 & 8192 & 16 & 0.0005 & 0.1 & 0.1 & 8 \\ \hline
iter2 (BT+ST) & \makecell[l]{Unigram \\ (16384)} & 378M & 10 & 1024 & 8192 & 16 & 0.001 & 0.2 & 0.2 & 64 \\ \hline
\end{tabular}
}
\caption{Hyper-parameters of the best performing \taen systems.}
\label{tab:hyperparameters_taen}
\end{table*}

\begin{table*}[t]
\small
\centering
\scalebox{0.85}{
\begin{tabular}{|l|l|l|l|l|l|l|l|l|l|l|}
\hline
System & \makecell[c]{Subword \\ (size)} & \# params & layers & \makecell[c]{embed \\ size} & \makecell[c]{ffn \\ embed size} & \makecell[c]{attention \\ heads} & \makecell[c]{learning \\ rate} & dropout & \makecell[c]{label \\ smoothing} & \# GPUs \\ \hline
\multicolumn{11}{|l|}{Constrained Tamil} \\ \hline
\makecell[l]{Baseline system \\ (bilingual \\ supervised)} & \makecell[c]{Unigram \\ (16384)} & 31M & 3 & 512 & 2048 & 8 & 0.0005 & 0.3 & 0.1 & 8 \\ \hline
iter1 (BT) & \makecell[c]{BPE \\ (20K)} & 314M & 10 & 1024 & 4096 & 16 & 0.0007 & 0.3 & 0.3 & 8 \\ \hline
iter2 (BT) & \makecell[c]{BPE \\ (20K)} & 314M & 10 & 1024 & 4096 & 16 & 0.0007 & 0.2 & 0.3 & 8 \\ \hline
\multicolumn{11}{|l|}{Unconstrained Tamil} \\ \hline
iter2 (BT) & \makecell[c]{BPE \\ (20K)} & 1.2B & 10 & 2048 & 8192 & 16 & 0.0001 & 0.3 & 0.1 & 8 \\ \hline
\end{tabular}
}
\caption{Hyper-parameters of the best performing \enta systems.}
\label{tab:hyperparameters_enta}
\end{table*}

\begin{table*}[t]
\small
\centering
\scalebox{0.85}{
\begin{tabular}{|l|l|l|l|l|l|l|l|l|l|l|}
\hline
System & \makecell[c]{Subword \\ (size)} & \# params & layers & \makecell[c]{embed \\ size} & \makecell[c]{ffn \\ embed size} & \makecell[c]{attention \\ heads} & \makecell[c]{learning \\ rate} & dropout & \makecell[c]{label \\ smoothing} & \# GPUs \\ \hline
\makecell[l]{Baseline system \\ (mBART+bi-FT)} & \makecell[c]{BPE \\ (130K)} & 487M & 12 & 1024 & 4096 & 16 & 3e-5 & 0.1 & 0.1 & 16 \\ \hline
\makecell[l]{NH-domain: \\ iter1-BT \\ (mBART+bi-FT)} & \makecell[c]{BPE \\ (130K)} & 487M & 12 & 1024 & 4096 & 16 & 1e-4 & 0.2 & 0.2 & 16 \\ \hline
\makecell[l]{news-domain: \\ iter1-BT} & \makecell[c]{BPE \\ (5K)} & 559M & 12 & 1024 & 8192 & 16 & 0.001 & 0.2 & 0.2 & 64 \\ \hline
\end{tabular}
}
\caption{Hyper-parameters of the best performing \iuen systems.}
\label{tab:hyperparameters_iuen}
\end{table*}

\begin{table*}[t]
\small
\centering
\scalebox{0.85}{
\begin{tabular}{|l|l|l|l|l|l|l|l|l|l|l|}
\hline
System & \makecell[c]{Subword \\ (size)} & \# params & layers & \makecell[c]{embed \\ size} & \makecell[c]{ffn \\ embed size} & \makecell[c]{attention \\ heads} & \makecell[c]{learning \\ rate} & dropout & \makecell[c]{label \\ smoothing} & \# GPUs \\ \hline
\makecell[l]{Baseline system \\ (bilingual \\ supervised)} & \makecell[c]{BPE \\ (5K)} & 122M & 4 & 1024 & 4096 & 8 & 0.001 & 0.3 & 0.3 & 4 \\ \hline
\makecell[l]{NH-domain: \\ iter1-ST} & \makecell[c]{BPE \\ (5K)} & 152M & 5 & 1024 & 4096 & 16 & 0.0005 & 0.2 & 0.2 & 4 \\ \hline
\makecell[l]{news-domain: \\ iter1-BT} & \makecell[c]{BPE \\ (5K)} & 152M & 5 & 1024 & 4096 & 16 & 0.001 & 0.2 & 0.2 & 4 \\ \hline
\end{tabular}
}
\caption{Hyper-parameters of the best performing \eniu systems.}
\label{tab:hyperparameters_eniu}
\end{table*}

\end{document}